\documentclass[runningheads]{llncs}
\usepackage{graphicx}

\usepackage{comment}
\usepackage{amsmath,amssymb} 
\usepackage{color}
\usepackage{booktabs}
\usepackage{multirow}
\usepackage{hyperref}

\usepackage[accsupp]{axessibility}  

\newcommand{\Ourds}{Transparent-460~}

\begin{document}
\pagestyle{headings}
\mainmatter

\title{TransMatting: Enhancing Transparent Objects Matting with Transformers} 


\titlerunning{TransMatting}
%
\author{
Huanqia~Cai$^*$ \and
Fanglei~Xue$^*$ \and
Lele~Xu$^{**}$ \and
Lili~Guo
}
\renewcommand{\thefootnote}{\fnsymbol{footnote}}
\footnotetext[1]{The first two authors contributed equally.}
\footnotetext[2]{Corresponding author. This work was supported by the National Natural Science Foundation of China under Grant 61901454. }
\footnotetext[0]{Project page: \url{https://github.com/AceCHQ/TransMatting}}

\renewcommand{\thefootnote}{\arabic{footnote}}

\authorrunning{Huanqia Cai et al.}
%
\institute{University of Chinese Academy of Sciences, Beijing, China \\
Key Laboratory of Space Utilization, Technology and Engineering Center \\ for Space Utilization, Chinese Academy of Sciences, Beijing, China\\
\email{\{caihuanqia19,xuefanglei19\}@mails.ucas.ac.cn \{xulele,guolili\}@csu.ac.cn }}
\maketitle

\begin{abstract}
Image matting refers to predicting the alpha values of unknown foreground areas from natural images. Prior methods have focused on propagating alpha values from known to unknown regions. However, not all natural images have a specifically known foreground. Images of transparent objects, like glass, smoke, web, etc., have less or no known foreground. In this paper, we propose a Transformer-based network, TransMatting, to model transparent objects with a big receptive field. Specifically, we redesign the trimap as three learnable tri-tokens for introducing advanced semantic features into the self-attention mechanism. A small convolutional network is proposed to utilize the global feature and non-background mask to guide the multi-scale feature propagation from encoder to decoder for maintaining the contexture of transparent objects. In addition, we create a high-resolution matting dataset of transparent objects with small known foreground areas. Experiments on several matting benchmarks demonstrate the superiority of our proposed method over the current state-of-the-art methods.

\end{abstract}

\section{Introduction}

Image matting is a technique to separate the foreground object and the background from an image by predicting a precise alpha matte as a result. It has been widely used in many applications, such as image and video editing, background replacement, and virtual reality \cite{chen2013knn,xu2017deep,levin2007closed}. Image matting assumes that every pixel in the image $I$ is a linear combination of the foreground object $F$ and the background $B$ by an alpha matte $\alpha$:

\begin{equation}
    I = \alpha F + (1-\alpha)B,   \alpha \in [0, 1]
\end{equation}

As only the image $I$ is known in this equation, the image matting is an ill-posed problem. So many existing methods \cite{chen2013knn,levin2007closed,sun2004poisson,wang2007optimized,xu2017deep,lu2019indices,li2020natural} take a trimap as an auxiliary input. The trimap segments the image into three parts: known foreground and background, and unknown area, indicated as white, black, and gray, separately.

Most traditional methods, including sampling-based \cite{berman2000method,chuang2001bayesian,wang2007optimized,gastal2010shared,he2011global,shahrian2013improving} and propagation-based methods \cite{chen2013knn,lee2011nonlocal,sun2004poisson,levin2007closed}, utilize the known area samples to find candidate colors or propagate the known alpha value. They heavily rely on the information from known areas, especially the known foreground areas. Recently, learning-based methods directly predict alpha mattes by neural network learning from well-annotated datasets. Although these methods take a great improvement in image matting, they also need specific information from known areas to predict unknown areas. However, according to \cite{liu2021long}, more than 50\% pixels in the unknown areas cannot be correlated to pixels in the known regions due to the limited reception field of deep learning methods. LFPNet \cite{liu2021long} proposes a Center-Surround Pyramid Pooling module to propagate the context feature from the known regions to the near unknown regions.
However, not all natural images have a salient and opaque object as the known foreground \cite{li2021deep}. Images of the glass, bonfires, plastic bags, etc., have salient foregrounds but with transparent or meticulous interiors; images of the web, smoke, water drops, etc., have non-salient foregrounds. The corresponding trimaps of these kinds of images will have very few or even no foreground areas. Most of the areas will be divided into the unknown regions. It is very challenging for existing models to learn long-range features with little known information.
Furthermore, with the development of modern cameras, picture resolution is becoming higher and higher. However, the reception fields of existing models could not increase as the resolution of input images do, which makes the problem even worse. 

To address this issue, we make the first attempt to introduce Vision Transformer (ViT) \cite{dosovitskiy2020image} to extract features with a large receptive field. The Transformer model is first proposed in natural language processing (NLP) and has achieved great performance in computer vision tasks, such as classification \cite{dosovitskiy2020image,touvron2021training,liu2021swin}, segmentation \cite{zheng2021RethinkingSemantic,lu2021SimplerBetter}, and detection \cite{carion2020DETR,yang2021focal}. It mainly consists of a multi-head self-attention (MHSA) module and a multi-layer perception module. The MHSA module could mine information in a global scope. Thus, the ViT model could learn global semantic features of the foreground object with high-level position relevance. To further help the model integrate the low-level appearance features (\textit{e.g.,} texture) with high-level semantic features (\textit{e.g.,} shape), a Multi-scale Global-guided Fusion (MGF) module is proposed. The MGF takes three adjacent scales of features as input, uses the non-background mask to guide the low-level feature, and employs the high-level feature to guide the information integration. With this new MGF module, only foreground features could be transmitted to the decoder, reducing the influence of background noises. 

Since the DIM \cite{xu2017deep} concatenates the trimap and RGB image to feed into the network, almost all subsequent trimap-based methods follow this strategy. However, compared with the RGB image, the trimap is very sparse and has some high-level positional relevance \cite{liu2021tripartite}. Most areas in the trimap have the same value, making convolution neural networks with small kernels inefficient in extracting features. Inspired by the \texttt{[cls]} token in ViT, we propose a new form of trimap named the tri-token map. Three learnable tokens are used to indicate the foreground, background, and unknown categories. We denote them as tri-tokens. Based on these tri-tokens, we propose a Tri-token Guided Transformer Block (TGTB), which adds the query with the corresponding tri-tokens for introducing the trimap information into the self-attention mechanism. With this high-level position information, the Transformer module could identify which features are from the known areas and which are from the unknown areas.

Besides, there has not been any testbed for images with transparent or non-salient foreground objects. Previous datasets mainly focus on salient and opaque foregrounds, like animals \cite{li2020endtoend} and portraits \cite{shen2016deep,liu2021tripartite}, which have significantly been investigated. To further help the community to dig into the transparent and non-salient cases, we collect 460 high-solution natural images with large unknown areas and manually label their alpha mattes.

Our main contributions can be summarized as follows:
\begin{enumerate}
\item We propose a TGTB module, introducing the Vision Transformer module to extract global semantic features with a big receptive field. We also redesign the trimap as a tri-token map to directly bring location information to the self-attention mechanism.

\item A MGF module is proposed to integrate multi-scale features, and the global information is well organized to guide the integration with low-level Transformer features.

\item We build a high-resolution matting dataset with 460 images of the transparent or non-salient foreground. The dataset will be released to promote the development of matting technology.

\item Experiments on three matting datasets demonstrate that the proposed TransMatting method outperforms the current SOTA methods, indicating the effectiveness of our proposed modules.

\end{enumerate}

\section{Related Works}
 In this section, we first briefly review matting from two perspectives: traditional methods and deep-learning methods. Then, we further give an overview of Vision Transformer models, as the Tri-token Guided Transformer Block (TGTB) is one of the main contributions of this work.

\subsection{Traditional Matting}
Traditional matting methods can be divided into two categories: sampling-based and propagation-based methods. These methods mainly rely on low-level features, like color, location, etc. The sampling-based methods \cite{berman2000method,chuang2001bayesian,wang2007optimized,gastal2010shared,he2011global,shahrian2013improving} first predict the colors of the foreground and background by evaluating the similarity of colors between the known foreground, background, and unknown area in samples, and then predict alpha mattes. Various sampling techniques have been investigated, including color cluster sampling \cite{shahrian2013improving}, edge sampling \cite{he2011global}, ray casting \cite{gastal2010shared}, etc. The propagation-based methods \cite{chen2013knn,lee2011nonlocal,sun2004poisson} propagate the information from the known foreground and background to the unknown area by solving the sparse linear equation system \cite{levin2007closed}, the Poisson equation system \cite{grady2005random}, etc., to obtain the best global optimal alpha.

\subsection{Deep-Learning Matting}
In recent decades, deep learning technologies have boomed in various fields of computer vision. The same goes for the image matting task. \cite{tang2019learning} combines the sampling and deep neural network to improve the accuracy of alpha matting prediction. The Indices matter method \cite{lu2019indices} proposes an index-guided method for up-sampling and down-sampling to make the detailed information in the prediction graph more complete. Based on providing a larger dataset Composition-1k \cite{xu2017deep}, DIM utilizes an encoder-decoder model to directly predict alpha mattes, which effectively improves the accuracy. \cite{sun2021semantic} introduces semantic classification information of the matting region and uses learnable weights and multi-class discriminators to revise the prediction results. \cite{yu2021mask} proposes a general matting framework, which is conducive to obtaining better results under the guidance of different qualities and forms. \cite{liu2021tripartite} further mines the information of the RGB map and trimap and fuses the global information from these maps for obtaining better alpha mattes. 
All of the above methods use trimap as guidance. Some trimap-free methods can predict alpha mattes without using trimap. However, the accuracy of these trimap-free methods still has a big gap compared to that of the trimap-guided ones \cite{chen2018semantic,yang2018active,qiao2020multi,yang2020smart}, indicating that the trimap could help the model to capture information efficiently.

\subsection{Vision Transformer}
The Transformer is firstly proposed in \cite{vaswani2017attention} to model long-range dependencies for machine translation and has demonstrated impressive performance on NLP tasks. Inspired by this, numerous attempts have been made to adapt transformers for vision tasks, and promising results have been shown for vision fields such as image classification, objection detection, semantic segmentation, etc. In particular, ViT \cite{dosovitskiy2020image} divides the input image into patches with a size of 16 $\times$ 16 and feeds the patch sequences to the vanilla Transformer model. To help the training process and improve the performance, DeiT \cite{touvron2021training} proposes a teacher-student strategy, which includes a distillation token for the student to learn from the teacher. Later, Swin \cite{liu2021swin}, PVT \cite{wang2021pyramid}, Crossformer \cite{wang2021crossformer}, and HVT \cite{pan2021scalable} combine the Transformer and pyramidal structure to decrease the number of patches progressively for obtaining multi-scale feature maps. To reduce computing and memory complexity, Swin, HRFormer \cite{yuan2021hrformer}, and CrossFormer apply local-window self-attention in Transformer, which also shows superior or comparable performance compared to the counterpart CNNs. The powerful self-attention mechanism in Transformer shows great advantages over CNN by capturing global attention of the whole image. However, some researchers \cite{li2021localvit} argue that locality and globality are both essential for vision tasks. Therefore, various researchers have tried combining the locality of CNN with the globality of Transformer to improve performance further. LocalViT \cite{li2021localvit} brings depth-wise convolutions to vision transformer to combine self-attention mechanism with locality, and shows great improvement compared to the pure Transformer, like DeiT, PVT, and TNT \cite{han2021transformer}.

\section{Matting Dataset}

According to the transparency of foregrounds, we could divide the images of matting into two types: 1) Transparent partially (TP): TP refers to that there are significant foreground and uncertainty areas, and the foreground areas can provide information for the prediction of uncertainty areas. For example, when the foreground is human, the opaque and unknown regions are the hair or clothes. 2)  Transparent totally (TT): there are minor or non-salient foreground areas, and the entire image is semi-transparent or high transparent. These images include glass, plastic bags, fog, water drops, etc.

As illustrated in Tab.~\ref{tab:dataset}, we select four popular image matting datasets for comparison, including DAPM \cite{shen2016deep}, Composition-1k, Distinctions-646 \cite{qiao2020attention}, and AIM-500 \cite{li2021deep}. The DAPM dataset consists only of portraits with no translucent or transparent objects. The Composition-1k dataset contains multiple categories, while most images are portraits (227 out of 481, TP-type). The Distinctions-646 dataset also mainly consists of portraits (343 out of 646, TP-type) \cite{liu2021tripartite}. The AIM-500 dataset contains only 76 TT-type images (correspond to the Salient Transparent/Meticulous type and the Non-Salient type in the original dataset) but 424 TP-type images.

\begin{table}[t]
\centering
\caption{Comparison between different public matting datasets. }
\label{tab:dataset}
\begin{tabular}{cccc} \toprule
Image Matting Dataset      & total num & TT num & resolution \\ \midrule
DAPM \cite{shen2016deep}           & 2000  & 0       & 800$\times$600    \\
Composition-1k \cite{xu2017deep}  & 481   & 86      & 1297$\times$1082  \\
Distinction-646 \cite{qiao2020attention} & 646   & 79      & 1727$\times$1565  \\
AIM-500 \cite{li2021deep}         & 500   & 76      & 1260$\times$1397  \\
\Ourds (Ours)                       & 460   & 460     & 3820$\times$3766  \\ \bottomrule
\end{tabular}
\end{table}

As we can see, the transparent objects in the above datasets only occupy a small portion. This may be because it is much more difficult to label transparent objects than other objects, limiting the progress of transparent objects in the matting field. In this work, we propose the first large-scale dataset targeting various high transparent objects called \Ourds dataset. Our \Ourds dataset includes 460 high-quality manually-annotated alpha mattes, where 410 images are for training and 50 for testing. Furthermore, to our best knowledge, the resolution of our \Ourds is the highest (the average resolution is up to 3820 $\times$ 3766) among all datasets with high transparent objects. We believe this new matting dataset will greatly advance the matting research on objects with massive transparent areas. 

\section{Methodology}

\begin{figure}[t]
\centering
\includegraphics[width=0.99\textwidth]{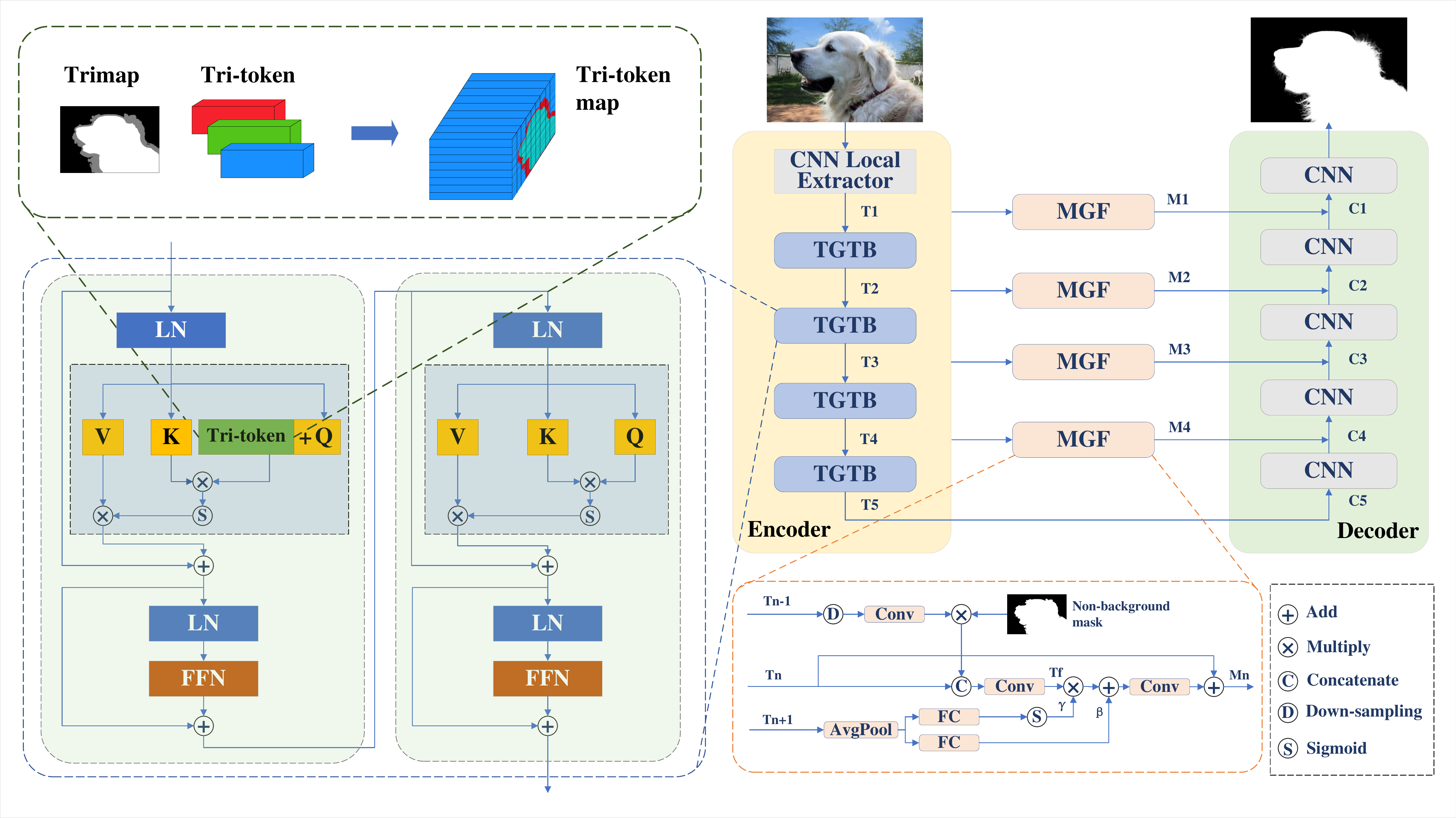}
\caption{The structure of our TransMatting.}
\label{fig:model}
\end{figure}

\subsection{Motivation}
By evaluating the results of some SOTA methods on TT and TP objects separately on the Composition-1k dataset (Tab.~\ref{tab:TTTP}), we find that the results of TT, which denotes the total transparent objects, are much worse than TP, indicating that TT objects are the key to affecting the overall evaluation results. Furthermore, we find that most of the existing methods rely on the information of the foreground region for predicting the unknown region \cite{chen2013knn,lee2011nonlocal,levin2008spectral,li2020natural,aksoy2017information}. However, such methods will become useless or ineffective when facing images with no definite known regions. For example, \cite{li2020natural} borrows features from both the known and unknown regions; when the unknown region is overwhelming in the images, the opacity propagation and the mattes prediction will face difficulties. Therefore, global information with a large or global receptive field and local features with inherent representation are needed to enhance the understanding and recognition capacity for objects with totally unknown regions. Although we can stack CNN layers to enlarge the receptive field, the information that covers the whole image is still hard to be obtained \cite{liu2021long}. Besides, CNN also lacks global connectivity \cite{li2021localvit}. By contrast, the Transformer is good at modeling long-range connectivity with its attention mechanism. 

\begin{table}[h]
\footnotesize
\centering
\caption{Performance of TT and TP objects on the Composition-1k dataset.}
\label{tab:TTTP}
\begin{tabular}{ccccccc}
\toprule
\multirow{2}{*}{Methods} & \multicolumn{3}{c}{MSE$\downarrow$} & \multicolumn{3}{c}{SAD$\downarrow$} \\ \cline{2-7} 
                         & TT     & TP    & TT+TP  & TT     & TP     & TT+TP \\ \hline
IndexNet \cite{lu2019indices}& 22.87  & 8.9   & 13     & 110.3  & 18.08  & 45.8  \\
GCAMatting \cite{li2020natural}& 15.89  & 6.2   & 9.1    & 85.72  & 13.68  & 35.3  \\
MGMatting \cite{yu2021mask}  & 13.01  & 4.65  & 7.18   & 77.88  & 11.87  & 31.76 \\ \midrule
TransMatting(Ours)                     & \textbf{7.49}   & \textbf{3.4}   & \textbf{4.58}   & \textbf{59.37}  & \textbf{10.35}  & \textbf{24.96} \\ \bottomrule
\end{tabular}
\end{table}

Moreover, most existing SOTA trimap-guided methods directly concatenate the trimap with RGB image as input. However, the huge gap between the two modalities of RGB image and trimap brings great difficulties in semantic feature extraction. At the same time, the trimap cannot effectively help the model focus on the region of interest. Therefore, a more efficient way to promote the guiding role of trimap is needed. 

In short, to improve the performance of TT objects, more global and local features should be captured, and an effective guidance method for the trimap should be developed.

\subsection{Baseline Structure}

To extract both the local and global features, we combine CNN and the Transformer model as our encoder. Specifically, the first part, like \cite{yu2021mask,li2020natural} is the same as the first two stages of ResNet34-UNet (denoted as CNN Local Extractor in Fig.~\ref{fig:model}). The second part consists of a stack of our proposed Tri-token Guided Transformer Block (TGTB) based on the Swin Transformer \cite{liu2021swin}. As the decoder, we adopt the original ResNet34-UNet, a widespread network in the matting field \cite{yu2021mask,li2020natural}.

\subsection{Trimap Guided Methods}
Almost all SOTA methods \cite{liu2021tripartite,yu2021mask,li2020natural,xu2017deep,yu2020high,sun2021semantic,cai2019disentangled} use trimap as a guide and directly concatenate the RGB image and the annotated trimap as the model's input. However, the modalities of the RGB image and trimap are quite different. The RGB image scales from 0 to 255 and shows fine low-level features like texture, color similarity, etc. The trimap includes three values, containing high-level semantic information, like shape, location, etc., \cite{liu2021tripartite}. Thus, the direct concatenation between them is not the most efficient way to extract features. 

Although trimap can explicitly indicate the region of interest, it is still hard to take full advantage of this information. To the best of our knowledge, we are the first to attempt to harmonize the RGB image and trimap rather than simply concatenating them.
We insert a learnable trimap into the Transformer module to guide the model to concentrate on the valuable area, making the network learning more efficient and robust.

\subsection{Tri-token}
Inspired by the \texttt{[cls]} token in Vision Transformer, we design a new tri-token (shown in Fig.~\ref{fig:model}) structure, aiming to introduce the high-level semantic information directly into the self-attention mechanism to replace the inefficient concatenation methods. 
Given a vanilla $Trimap \in \mathbb{R}^{H \times W}$, we generate three \textbf{learnable} tri-tokens (denoted as $Token_i$, $i=\{0,1,2\}$) with different initialization to represent the known foreground, known background, and unknown areas, respectively. Every tri-token is a 1D vector, that is, $Token_i \in \mathbb{R}^{C}$.
Then we replace every pixel in the trimap with the corresponding tri-token to generate the tri-token map, formulated as:
\begin{equation}
    Trimap[Trimap==i] = Token_i,   i=\{0,1,2\}
\end{equation}
In this manner, the tri-token map can directly guide the self-attention process in the Transformer to pay more attention to the unknown areas for self-updating.

\subsection{Tri-token Guided Transformer Block}  
Global connectivity is much more important for the prediction of total transparent objects. CNN does not have global attention, and its receptive field cannot cover the whole image \cite{liu2021long}, which leads to poor estimation for pixels outside receptive fields, while Transformer has global attention, and its receptive field can cover every pixel at the first layer. 

The Transformer consists of multi-head self-attention (MHSA) and Multilayer Perceptron (MLP) blocks. The self-attention mechanism can be thought of as a mapping between a query and a collection of key-value pairs. The output is a weighted sum of the values, and the weights are assigned by the compatibility function between the query and the relevant key. This can be implemented by Scaled Dot-Product Attention \cite{vaswani2017attention}, in which a softmax function is used to activate the dot products of query and all keys for obtaining the weights. MHSA means that more than one self-attention is performed in parallel.

Like \cite{liu2021swin,wang2021crossformer,yuan2021hrformer}, we use non-overlapping windows whose size is $M \times M$ to divide the feature maps. The MHSA is performed within each window. The formulations of vanilla attention and our tri-token attention in a specific window are shown as follows:
\begin{equation}
    Attention(Q, K, V) = Softmax(QK^T / \sqrt{d})V
\end{equation}

\begin{equation}
    Tri\mbox{-}token\ Attention(Q, K, V) = Softmax((Q + Tri\mbox{-}token)K^T / \sqrt{d})V
\end{equation}

where $Q, K, V \in R^{M^2 \times d}$ represent the query, key, and value in the attention mechanism, respectively. $d$ is the query/key dimension. In the Tri-token Attention formulation, $Q$, $K$, and $V$ are the same as that in the standard self-attention. The $Tri\mbox{-}token$ is our proposed learnable \emph{trimap} that adds to the query for forming a new tri-token query. In this way, our tri-token attention mechanism can selectively aggregate contexts and evaluate which region should be paid more attention to with the guidance of our learnable tri-tokens.

In this way, we combine the self-attention and tri-tokens to focus on more valuable regions by considering the relationship between non-background and background areas, and finally achieve the best performance. We use our tri-token attention every five blocks in each Tri-token Guided Transformer Block (TGTB).

\subsection{Multi-scale Global-guided Fusion Module}

In the multi-scale feature pyramid structure, in-depth features contain more global information, while shallow features have rich local information like texture, color similarity, etc. Fusing these features is vital for accurately predicting alpha mattes for high transparent objects \cite{qiao2020multi}.
Although the direct sum operation can realize feature fusion, the details in the shallow features may attenuate the impact of the advanced semantics, resulting in some subtle regions missing \cite{qiao2020multi}. To address this issue, we propose a Multi-scale Global-guided Fusion (MGF) module in the decoder process (see Fig.~\ref{fig:model} for details), with both the non-background information and the advanced semantic features as guidance, to fuse the high-level semantic information and the lower ones effectively.

Specifically, we denote three adjacent features from shallow to deep as $T_{n-1}$, $T_{n}$, and $T_{n+1}$. The $T_{n-1}$ is first down-sampled, then the Hadamard product is employed between the non-background mask and $T_{n-1}$ to extract the low-level features of non-background, which helps to reduce the impact of complex background influence. This can guide the network to pay more attention to the foreground and unknown areas. After that, the $T_{n-1}$ is concatenated with $T_{n}$, and a convolution layer is performed to align the channel of fused features. We mark this feature as $T_{f}$. 

For the $T_{n+1}$, we first perform a global average pooling to generate channel-wise statistics and then use two fully connected (FC) layers to squeeze channels. As shown in Fig.~\ref{fig:model}, features output from the two FC layers are denoted as $\gamma$ and $\beta$, separately.To fully capture channel-wise dependencies, we add a sigmoid function to activate $\gamma$ and perform broadcast multiplication with $T_{f}$ for channel re-weighting. After that, broadcast addition is performed between the channel-weighted feature and $\beta$. A convolution layer is used to fuse information from different groups. Notably, a skip connection from $T_{n}$ is employed for obtaining the final fused features of MGF.

In short, considering that fusing low-level features directly may cause a negative impact on the advanced semantics \cite{qiao2020multi}, two techniques are proposed here. Firstly, the non-background mask is introduced into the fusion process to filter out the complex background information and further help to concentrate more attention on the foreground and unknown areas. Secondly, the global channel-wise attention from higher-level features is used for re-weighting and enhancing the important information in the fused features.

\subsection{Loss Function}
Following \cite{yu2021mask}, we use three losses, including the alpha loss ($\mathcal{L}_\alpha$), Compositional loss \cite{xu2017deep} ($\mathcal{L}_{comp}$), and Laplacian loss \cite{hou2019context} ($\mathcal{L}_{lap}$). As formulated below, their weights are set as 0.4, 1.2, and 0.16, respectively.

\begin{equation}
    \mathcal{L}_{final} = 0.4*\mathcal{L}_{\alpha} + 1.2*\mathcal{L}_{comp} + 0.16*\mathcal{L}_{lap}
\end{equation}

\section{Experiments}
In this section, we show our experimental settings and compare our evaluation results on the test set of Composition-1k \cite{xu2017deep}, Distinction-646 \cite{qiao2020attention}, and our \Ourds datasets with other state-of-the-art methods.

\subsection{Dataset}

\textbf{Composition-1k} contains 431 and 50 unique foreground objects and manually labeled alpha mattes as training and test sets, respectively. Every foreground object is composited with 100 (for training set) and 20 (for test set) background images from COCO \cite{lin2014microsoft} and Pascal VOC \cite{everingham2010pascal}. As a result, there are 43,100 images for training and 1,000 images for testing.

\textbf{Distinction-646} comprises 646 distinct foreground objects. Similar to the Composition-1k, 50 objects are divided as the test set. Following the same composition rule, there are 59,600 and 1000 images for training and testing, respectively.

\textbf{Our \Ourds} mainly consists of transparent and non-salient objects as the foreground, like water drops, jellyfish, plastic bags, glass, crystals, etc. We collect 460 high-resolution images and carefully annotate them with Photoshop. Considering the transparent objects are very meticulous, we keep the original resolution of all collected images (up to 3820 $\times$ 3766 pixels on average). To our best knowledge, this is the first transparent object matting dataset in such a high resolution.

\subsection{Evaluation Metrics}
Following \cite{hou2019context,cai2019disentangled,lu2019indices,liu2021tripartite}, we use four metrics for evaluation, including the Sum of Absolute Differences (SAD), Mean Squared Error (MSE), Gradient error (Grad.) and Connectivity error (Conn). It is notable that the unit of MSE value is set to 1e-3 for easy reading.

\subsection{Implementation Details}
We use PyTorch \cite{paszke2019pytorch} to implement our proposed method. All the experiments are trained for 200,000 iterations. We initialize our network with ImageNet \cite{deng2009imagenet} pre-trained weights. The ablation experiments in Tab.~\ref{tab:ab:main}, 4, \ref{ab:main} are done with 2 NVIDIA Tesla V100 GPU with a batch size of 32. Moreover, to compare our method with the existing SOTA methods, we use a batch size of 64 with 4 NVIDIA Tesla V100 GPU to train our proposed method in Tab.~\ref{tab:sota:AIM}, \ref{tab:sota:Distinctions}, \ref{tab:sota:ours}. The Adam optimizer is utilized, and the initial learning rate is set to 1e-4 with the same learning rate decay strategy as \cite{yu2021mask,loshchilov2016sgdr}. For a fair comparison, we follow the data augmentation methods used in \cite{li2020natural}, like random crop, rotation, scaling, shearing, etc. Moreover, the trimaps for training are generated using dilation and erosion ways on alpha images by random kernel sizes from 1 to 30. Finally, we crop 512×512 patches on the center of the unknown area of alpha and composite them with the background from COCO. We use the same training conditions on the Composition-1k and Distinction-646 datasets.

\subsection{Ablation Study}

To evaluate the effectiveness of our new proposed modules of TGTB and MGF, and the performance with different hyper-parameters, we design the ablation study on the Composition-1k dataset.

\textbf{Evaluate the effectiveness of our proposed modules.} The quantitative results under the SAD, MSE, Gradient, and Connectivity errors with and without our proposed TGTB and MGF modules are illustrated in Tab.~\ref{tab:ab:main}. As we can see, with the TGTB module, the four metrics listed above decrease to 27.45, 5.66, 11.77, and 24.30, respectively. The main reason is that our redesigned tri-token map is more suitable for propagating location information than simply concatenating to the input image. The MGF module could solely achieve similar performance, indicating that our proposed multi-scale feature fusion strategy can also help the decoder to make better use of the local and global information. When combined with the TGTB and MGF modules, the model achieves the best performance, indicating the effectiveness of the two new proposed modules.

\begin{table}[t]
\centering
\caption{The effectiveness of our proposed TGTB and MGF modules on the Composition-1k dataset.}
\label{tab:ab:main}
\begin{tabular}{@{}cccccc@{}}
\toprule
TGTB   &   MGF        & SAD $\downarrow$   & MSE $\downarrow$  & Grad. $\downarrow$  & Conn.$\downarrow$  \\ \midrule
       &       & 29.14     & 6.34    & 12.06     & 25.21     \\
$\checkmark$ & & 27.45     & 5.66   & 11.77     & 24.30       \\ 
& $\checkmark$  & 27.21     & 5.57    & 11.23     & 23.25     \\ 
$\checkmark$ & $\checkmark$ & \textbf{26.83} & \textbf{5.22} & \textbf{10.62} & \textbf{22.14} \\ \bottomrule
\end{tabular}
\end{table}

\textbf{Determine where to introduce tri-tokens.}
There are four TGTB stages in our encoder model. Tab.~4 reports the performance with different positions to introduce tri-tokens. As the position goes deep, the feature map size decreases, making more position information lose. On the other hand, deep stages have learned more abstract semantic features, which is suitable for mutual learning with tri-tokens. As shown in Tab.~4, both shallow and deep stages benefit from tri-tokens, indicating that the tri-tokens in TGTB modules could guide the encoder to focus on the right regions.

\begin{table}[t]
\begin{minipage}[t]{0.5\columnwidth}
    \centering
    \label{tab:ab:position}
    \parbox{0.96\columnwidth}{\caption{Ablation results on the Composition-1k dataset with different positions to introduce the proposed tri-tokens.}}
    \begin{tabular}{@{}ccccc@{}}
    \toprule
    Position        & SAD$\downarrow$   & MSE$\downarrow$  & Grad.$\downarrow$  & Conn.$\downarrow$  \\ \midrule

1     & 31.68 & 7.24 & 14.20 & 27.42  \\
4     & 29.50 & 6.20 & 13.18 & 25.23  \\
1,2,3,4       & \textbf{26.83} & \textbf{5.22} & \textbf{10.62} & \textbf{22.14} \\  \bottomrule
    \end{tabular}
\end{minipage}\begin{minipage}[t]{0.5\columnwidth}

\centering
\caption{Ablation results on the Composition-1k dataset with local or (and) global features in the proposed MGF module.}
\label{ab:main}
\begin{tabular}{@{}cccccc@{}}
\toprule
local & global        & SAD$\downarrow$   & MSE$\downarrow$  & Grad.$\downarrow$  & Conn.$\downarrow$  \\ \midrule
 &                & 27.45     & 5.66     & 11.77     & 24.30     \\
 $\checkmark$ &     & 27.16     & 5.34    & 11.03     & 22.60     \\ 
 & $\checkmark$      & 27.39     &  5.46   & 11.43      & 23.20     \\ 
                $\checkmark$&$\checkmark$   & \textbf{26.83} & \textbf{5.22} & \textbf{10.62} & \textbf{22.14} \\ \bottomrule
\end{tabular}
\end{minipage}
\end{table}

\textbf{The impact of local and global features in MGF.}
Tab.~5 reports the effectiveness of our MGF module with and without local or global branches. The local branch is proposed to integrate $T_{n-1}$ with the non-background mask, and the global branch is responsible for introducing global features from $T_{n+1}$ to guide the feature flow. As we can see from Tab.~5, combining local and global branches could achieve the best performance compared with using one of them solely. The main reason is the effectiveness of our MGF in fusing local (texture, border) and global (semantic, location) features for modeling unknown regions.

\begin{figure}[t]
\centering
\includegraphics[width=0.98\textwidth]{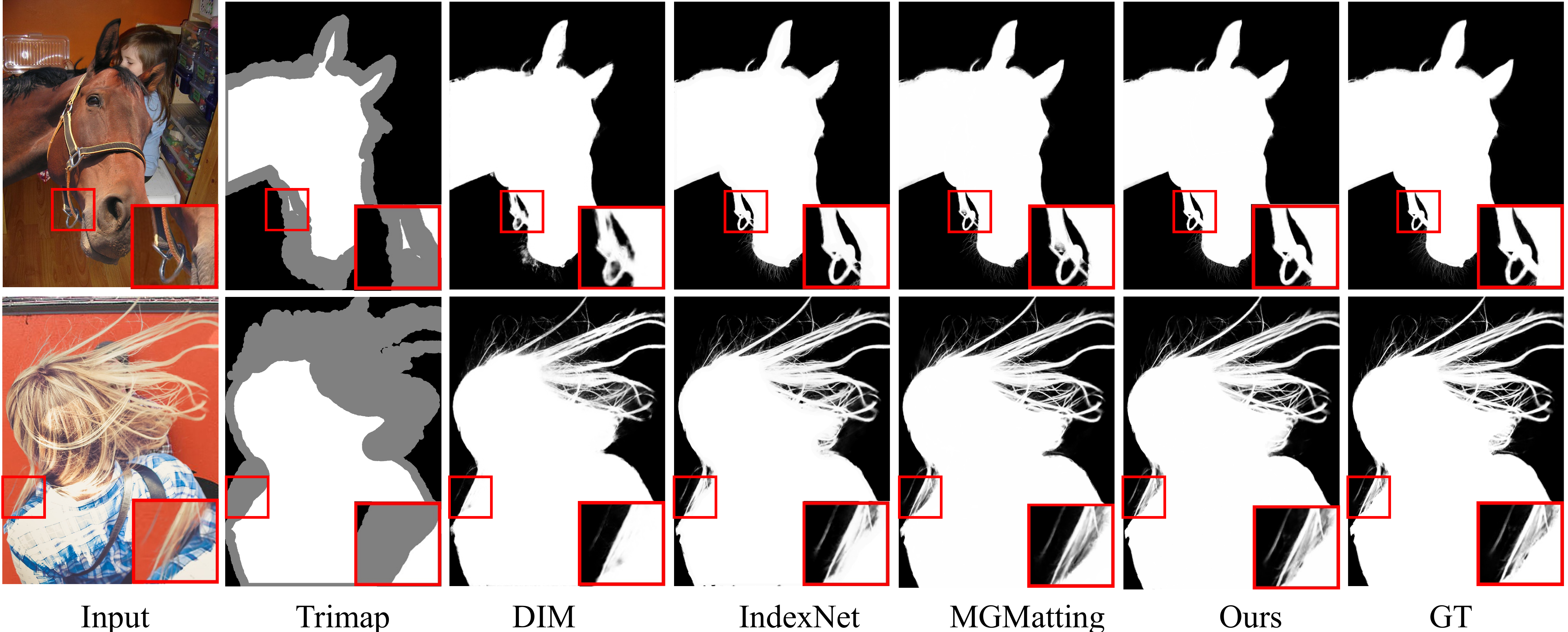}
\caption{Visual comparison of our TransMatting against SOTA methods on the Composition-1k test set.}
\label{fig:composition1K}
\end{figure}

\subsection{Comparison with Prior Work}
To evaluate our method's performance, we compare it with other state-of-the-art models on the following three datasets. Notably, we achieve the best performance on all three datasets.

\textbf{Testing on Composition-1k.}
We show the quantitative and visual results on Tab.~\ref{tab:sota:AIM} and Fig.~\ref{fig:composition1K}. Without any test-time augmentations, our proposed TransMatting outperforms other SOTA methods on all four evaluation metrics by only using the Composition-1k training set for training. As illustrated in Tab.~\ref{tab:sota:AIM}, our model decreases the MSE and Grad metrics heavily: from 5.2, 10.6 to 4.58 and 9.72, respectively, indicating the effectiveness of our TransMatting.

\begin{table}[t]
\caption{The quantitative results on the Composition-1k test set \cite{xu2017deep}.  $^{\dagger}$ denotes results with test-time augmentation.} 
\label{tab:sota:AIM}
\centering
\begin{tabular}{@{}c|cccc@{}}
\toprule
Methods & SAD$\downarrow$& MSE$\downarrow$ & Grad.$\downarrow$ & Conn.$\downarrow$ \\ \toprule
AlphaGAN \cite{lutz2018alphagan}      & 52.4  & 30  & 38   & 53    \\
DIM \cite{lu2019indices}           & 50.4  & 14  & 31.0 & 50.8  \\
IndexNet \cite{lu2019indices}      & 45.8  & 13  & 25.9 & 43.7  \\
AdaMatting \cite{cai2019disentangled}    & 41.7  & 10  & 16.8 & -     \\
ContextNet \cite{hou2019context}    & 35.8  & 8.2 & 17.3 & 33.2  \\
GCAMatting \cite{li2020natural}    & 35.3  & 9.1 & 16.9 & 32.5  \\
MGMatting \cite{yu2021mask}    & 31.5  & 6.8  & 13.5 & 27.3 \\
TIMI-Net \cite{liu2021tripartite}      & 29.08 & 6.0 & 12.9 & 27.29 \\
FBAMatting \cite{forte2020f} $^{\dagger}$     & 25.8  & 5.2 & 10.6 & 20.8  \\ \midrule
TransMatting(Ours) & \textbf{24.96}& \textbf{4.58} & \textbf{9.72} & \textbf{20.16}  \\ \bottomrule
\end{tabular}
\end{table}

\begin{figure}[t]
\centering
\includegraphics[width=0.99\textwidth]{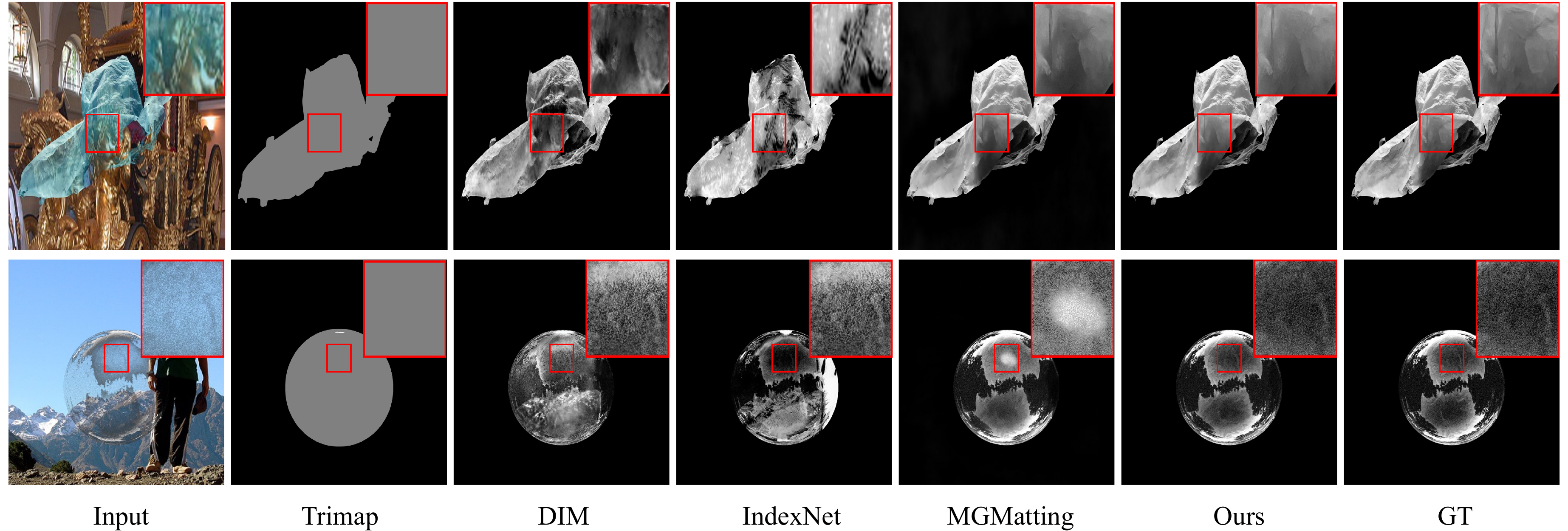}
\caption{Visual comparison of our TransMatting against SOTA methods on our \Ourds \ test set.}

\label{fig:composition}
\end{figure}

\textbf{Testing on Distinction-646.}
Tab.~\ref{tab:sota:Distinctions} compares the performance of our TransMatting with other state-of-the-art methods on Distinction-646. For a fair comparison, we follow the whole inference protocol in \cite{qiao2020attention,yu2021mask} to calculate the metrics based on the whole image. Without any additional tuning, our method outperforms all the SOTA methods.

\textbf{Testing on our \Ourds.}
Based on their release codes, we train IndexNet and MGMatting methods on our dataset and compare them with ours in Tab.~\ref{tab:sota:ours}. Our \Ourds dataset mainly focuses on transparent and non-salient foregrounds, which is very difficult for existing image matting methods. Surprisingly, as illustrated in Tab.~\ref{tab:sota:ours}, our TransMatting achieves promising results with only a 4.02 MSE error.
Furthermore, to evaluate the generalization performance of our model. We train our TransMatting on the Composition-1k training set and directly test it on the \Ourds test set. The results are shown in Tab.~\ref{tab:generalize}. Thanks to the big receptive field and well-designed multi-scale fusion module, our model reduces nearly half of the SAD, MSE, and Conn. errors compared to the SOTA methods.

\begin{table}[t]
\centering
\caption{The quantitative results on the Distinction-646 test set. }
\label{tab:sota:Distinctions}
\begin{tabular}{@{}c|cccc@{}}
\toprule
Methods       & SAD$\downarrow$   & MSE$\downarrow$  & Grad.$\downarrow$  & Conn.$\downarrow$  \\ \midrule
KNNMatting \cite{chen2013knn}    & 116.68 & 25   &103.15 & 121.45 \\
DIM \cite{xu2017deep}           & 47.56 & 9    & 43.29 & 55.90 \\
HAttMatting \cite{qiao2020attention}   & 48.98 & 9    & 41.57  & 49.93 \\
GCAMatting \cite{li2020natural}    & 27.43 & 4.8  & 18.7 & 21.86 \\ 
MGMatting \cite{yu2021mask} & 33.24 & 4.51 & 20.31 & 25.49 \\ \midrule
TransMatting (Ours) & \textbf{25.65} & \textbf{3.4}     & \textbf{16.08}      & \textbf{21.45}       \\ \bottomrule
\end{tabular}
\end{table}

\begin{table}[t]
\centering
\caption{The quantitative results on our proposed \Ourds test set.}
\label{tab:sota:ours}
\begin{tabular}{@{}c|cccc@{}}
\toprule
Methods       & SAD$\downarrow$   & MSE$\downarrow$ & Grad.$\downarrow$  & Conn.$\downarrow$  \\ \midrule
IndexNet \cite{lu2019indices}      & 573.09 & 112.53  & 140.76 & 327.97 \\
MGMatting \cite{yu2021mask}    & 111.92 & 6.33  & 25.67 & 103.81 \\
TransMatting (Ours) & \textbf{88.34}      & \textbf{4.02}    & \textbf{20.99}      & \textbf{82.56}     \\ \bottomrule
\end{tabular}
\end{table}

\begin{table}[h!]
\centering
\caption{The generalization results on our proposed \Ourds   test set.}
\label{tab:generalize}
\begin{tabular}{@{}c|cccc@{}}
\toprule
Methods       & SAD$\downarrow$   & MSE$\downarrow$ & Grad.$\downarrow$  & Conn.$\downarrow$  \\ \midrule
DIM \cite{xu2017deep}       & 356.2   & 49.68   & 146.46   & 296.31  \\
IndexNet \cite{lu2019indices}      & 434.14  & 74.73   & 124.98  & 368.48\\
MGMatting \cite{yu2021mask}    & 344.65  & 57.25   & 74.54    & 282.79\\
TIMI-Net \cite{liu2021tripartite}  & 328.08      &  44.2    & 142.11     & 289.79      \\ 
TransMatting(Ours) & \textbf{192.36}      & \textbf{20.96}    &  \textbf{41.8}     & \textbf{158.37}      \\ \bottomrule
\end{tabular}
\end{table}

\section{Conclusion}

In order to generalize to transparent and non-salient foregrounds, matting algorithms must have the ability to mine long-range features and utilize the semantic features in trimap. In this paper, we propose a novel Transformer-based network by redesigning a tri-token map to introduce the trimap semantic features into the long-range dependencies of the self-attention mechanism. Furthermore, a multi-scale global-guided fusion module is proposed to take the global information and local non-background mask as a guide to fuse multi-scale features for better modeling the unknown regions in transparent objects. Experiments on the Composition-1k, Distinctions-646, and our proposed \Ourds datasets demonstrate that our TransMatting outperforms the state-of-the-art methods.

\clearpage
%
%
\bibliographystyle{splncs04}
\bibliography{egbib}
\end{document}